# Real-Time Surgical Instrument Defect Detection via Non-Destructive Testing

Qurrat Ul Ain[1*], Atif Aftab Ahmed Jilani[2], Zunaira Shafqat[3], Nigar Azhar Butt[4]

[1,2,3*]Department of Software Engineering, National University of Computer and Emerging Science, A.k Brohi Rd, Islamabad, 46000, Pakistan.
.

*Corresponding author(s). E-mail(s): quratul.ain.v@isb.nu.edu.pk;
Contributing authors: atif.jilani@nu.edu.pk; zunairahamza7@gmail.com; nigar.azhar@isb.nu.edu.pk;

These authors contributed equally to this work.

**Abstract**

Defective surgical instruments pose serious risks to sterility, mechanical integrity, and patient safety, increasing the likelihood of surgical complications. However, quality control in surgical instrument manufacturing often relies on manual inspection, which is prone to human error and inconsistency. This study introduces SurgScan, an AI-powered defect detection framework for surgical instruments. Using YOLOv8, SurgScan classifies defects in real-time, ensuring high accuracy and industrial scalability. The model is trained on a high-resolution dataset of 102,876 images, covering 11 instrument types and five major defect categories. Extensive evaluation against state-of-the-art CNN architectures confirms that SurgScan achieves the highest accuracy (99.3%) with real-time inference speeds of 4.2–5.8 ms per image, making it suitable for industrial deployment. Statistical analysis demonstrates that contrast-enhanced preprocessing significantly improves defect detection, addressing key limitations in visual inspection. SurgScan provides a scalable, cost-effective AI solution for automated quality control, reducing reliance on manual inspection while ensuring compliance with ISO 13485 and FDA standards, paving the way for enhanced defect detection in medical manufacturing.

# 1 Introduction

The reliability and safety of surgical instruments directly influence patient outcomes and procedural success. Defects such as cracks, corrosion, scratches, or structural misalignments significantly raise the risk of surgical-site infections and complications, posing critical safety concerns [1]-[2]. Despite rigorous international standards, quality control in surgical instrument manufacturing often relies heavily on manual visual inspection, a method inherently subjective, inconsistent, labor-intensive, and prone to human errors, especially in detecting subtle defects like micro-cracks, tiny pores, and early-stage corrosion [3, 4].

For high-volume surgical instrument manufacturers, especially in key global exporting regions such as Pakistan, manual inspection limits scalability and raises the risk of export rejection, economic losses, and reputational damage due to undetected defects. This underscores the necessity for reliable, automated inspection techniques capable of accurately identifying subtle defects consistently and efficiently. [4].

To meet global regulatory standards, all surgical instruments must undergo rigorous inspection processes. Industry guidelines require instruments to be free from physical defects, including cracks, scratches, corrosion, and structural misalignments [5]. Additionally, medical standards recommend magnified inspection to detect microlevel debris and imperfections that might be invisible to the naked eye [6]. However, the subjectivity of manual inspections, coupled with high labor costs and time constraints, underscores the need for an automated, scalable, and efficient quality control solution.

Recent advancements in deep learning-based object detection have shown promising potential for addressing these inspection challenges. Traditional Convolutional Neural Networks (CNNs) have significantly advanced defect detection accuracy, yet achieving real-time performance, particularly for micro-level defects, remains a significant challenge. To bridge this gap, modern object detection architectures like You Only Look Once (YOLO) have emerged, demonstrating considerable promise in achieving real-time accuracy and industrial scalability.

To address these industry challenges, this paper introduces SurgScan, an AIpowered defect detection framework developed explicitly for automated, real-time inspection of surgical instruments. SurgScan leverages the advanced YOLOv8 architecture, optimized to effectively classify instrument types and detect manufacturing defects in real-time, with the potential for high accuracy and industrial deployment.

1. **Creation of a High-Resolution Surgical Defect Dataset** – We introduce a large-scale, expertly annotated image dataset covering multiple surgical instrument types and common defect categories, enriched through extensive data augmentation to enhance robustness and generalization.
2. **Implementation and Benchmarking of YOLOv8** – We systematically compare the performance of the YOLOv8 architecture against prominent CNN models, evaluating their potential effectiveness for real-time surgical instrument inspection in industrial settings.
3. **Statistical Validation of Augmentation and Preprocessing Methods** – We statistically examine the impact of data augmentation strategies and preprocessing adjustments (contrast, brightness, sharpness) on the defect classification accuracy and balance of defect distribution, offering evidence-based insights for optimal dataset preparation.

4. **Industrial Relevance and Practical Impact** – By automating defect detection, SurgScan ensures compliance with international medical quality standards, reduces reliance on manual inspection, minimizes human error, enhances production efficiency, and strengthens Pakistan's position in the global surgical instrument industry.

The paper is structured as follows: Section 2 provides essential background on quality control and regulatory requirements; Section 3 surveys related works in automated defect detection; Section 4 presents our detailed dataset curation process; Section 5 describes the SurgScan methodology; Section 6 clearly defines the research questions and experimental setup; Section 7 presents and analyzes experimental results; Section 8 discusses threats to validity; and Sections 9 and 10 summarize conclusions and outline future research directions, respectively.

## 2 Background

Ensuring the quality and reliability of surgical instruments is crucial for maintaining patient safety and preventing surgical complications. Defective instruments with cracks, corrosion, scratches, or misalignments can compromise sterility and mechanical integrity, increasing the risk of surgical-site infections and procedural failures [7][2]. To mitigate these risks, regulatory bodies such as the U.S. Food and Drug Administration (FDA) and the International Organization for Standardization (ISO) enforce strict quality control guidelines that require surgical instruments to be free from defects before clinical use [2]. Specifically, ISO 13485 mandates rigorous quality assurance processes for medical device manufacturing, while FDA-GMP (Good Manufacturing Practices) regulations outline inspection standards to ensure compliance [3]. Despite these stringent regulations, many manufacturers still struggle with efficient defect detection, particularly in high-volume production environments [4].

### 2.1 Limitations of Manual Visual Inspection

The most common approach to surgical instrument quality control is manual visual inspection, where trained inspectors examine instruments for surface defects under standard lighting conditions. However, this method suffers from several critical limitations [5]. First, human subjectivity plays a major role—inspectors may overlook subtle defects due to fatigue, lighting inconsistencies, or visual limitations [6]. Studies have shown that small scratches, micro-cracks, and early-stage corrosion are often missed during manual inspections, leading to potential product recalls or export rejections [8]. Second, scalability issues arise in high-volume manufacturing settings, where thousands of instruments must be inspected daily. Manual inspection is not only time-consuming but also costly, requiring extensive labor and slowing down production lines [9]. Finally, environmental factors such as lighting variations and reflective surfaces can further complicate defect identification, making it difficult to maintain consistency across inspections [10]. These challenges highlight the urgent need for automated, AI-driven quality control solutions that can enhance accuracy, efficiency, and reliability.

## 2.2 Transition to Automated Inspection: Deep Learning Approaches

To overcome the limitations of manual inspection, researchers and industry experts have explored automated defect detection using computer vision and deep learning techniques [11]. Traditional machine learning methods, such as template matching and edge detection, have been applied to identify defects in industrial settings. However, these approaches struggle with variability in instrument textures, lighting conditions, and defect appearances, leading to high false positive and false negative rates [12].

The advent of Convolutional Neural Networks (CNNs) has significantly improved defect detection accuracy by enabling automatic feature extraction from images [13]. CNN-based models, such as ResNet and EfficientNet, have demonstrated strong performance in industrial defect classification tasks [14]. However, these models typically require multiple forward passes and high computational power, making them less suitable for real-time quality control in manufacturing environments [15].

To address real-time processing constraints, YOLO-based object detection models have gained popularity for their ability to detect and classify defects in a single forward pass, significantly reducing inference time [16]. YOLO architectures, including YOLOv5 and YOLOv8, have demonstrated high-speed and high-accuracy performance, making them ideal for real-time defect detection in industrial applications [17]. However, despite their advantages, challenges remain in applying deep learning models to surgical instrument defect detection, particularly due to dataset limitations, reflective surfaces, and fine-grained defect classification [18].

## 3 Related Work

Over the years, defect detection in industrial and medical manufacturing has evolved from manual inspection to computer vision-based solutions. Traditional techniques, such as thresholding, edge detection, and morphological operations, have been widely used for detecting surface defects in metallic and industrial components [19]. While effective for basic defect localization, these methods struggle with variations in lighting, surface reflectivity, and defect shape complexity [20]. To overcome these limitations, machine learning and deep learning techniques have gained prominence in automated defect detection.

### 3.1 Traditional Image Processing-Based Approaches

Earlier research in defect detection relied on rule-based image processing techniques such as Canny edge detection, Hough transforms, and contour analysis [21]. These methods were applied in various industrial applications, including surface defect inspection for steel and electronic components. For example, several studies employed wavelet transforms and local binary patterns (LBP) to detect corrosion and scratches in metallic surfaces [22]. However, these approaches rely heavily on handcrafted feature extraction and often perform poorly under non-uniform lighting conditions and surface variations [23].

## 3.2 Machine Learning-Based Defect Detection

With advancements in computer vision, researchers have explored machine learningbased classifiers such as Support Vector Machines (SVM), Random Forests, and KNearest Neighbors (KNN) for defect classification [24]. These models extract textural, statistical, and shape-based features from defect images to train a supervised classifier. While these approaches demonstrated improvements over traditional methods, they are highly dependent on feature engineering, requiring domain expertise to manually design robust descriptors [25]. Additionally, conventional machine learning models lack the scalability required for real-time, large-scale quality control in industrial settings.

## 3.3 Deep Learning for Automated Defect Detection

The emergence of Convolutional Neural Networks (CNNs) revolutionized defect detection by enabling automated feature extraction directly from raw images [26]. CNN architectures, such as ResNet, DenseNet, and EfficientNet, have been applied to classify and segment surface defects with high accuracy [27]. Studies have shown that CNN-based methods outperform traditional techniques by learning hierarchical feature representations, making them robust to illumination changes, varying defect sizes, and complex textures [28]. However, CNN-based models typically require multiple forward passes and high computational power, making them less suitable for real-time manufacturing environments.

To address the need for faster and more efficient inference, researchers have turned to object detection models such as YOLO (You Only Look Once) and SSD (Single Shot MultiBox Detector) [29]. YOLO-based architectures, in particular, have demonstrated superior real-time performance by detecting defects in a single forward pass, significantly reducing latency compared to CNN classifiers [30]. Studies implementing YOLOv3 and YOLOv5 for industrial defect detection have reported high accuracy and inference speeds, making them promising candidates for automated quality control in manufacturing. However, existing works often struggle with fine-grained defect detection, particularly in highly reflective and metallic surfaces, such as surgical instruments
[31].

## 3.4 Limitations in Existing Research & Need for SurgScan

Despite significant progress in defect detection research, existing methods still face several key challenges when applied to surgical instrument inspection. Many deep learning models are trained on generic industrial datasets that lack high-resolution, well-annotated data specific to surgical instruments [32]. Additionally, most prior studies focus on defect classification rather than real-time detection and localization, making them less practical for high-volume, automated inspection pipelines. To bridge these gaps, this paper presents SurgScan, a real-time deep-learning framework designed specifically for automated defect detection in surgical instruments.

# 4 Dataset Curation

Existing research highlights a critical challenge in deep learning-based surgical instrument defect detection: the limited availability of comprehensive and well-annotated datasets.While synthetic datasets, such as those generated using 3D Gaussian splatting [33], provide a controlled environment for defect modeling, they fail to replicate real-world inconsistencies in material texture, lighting variations, and surface wear. These factors play a crucial role in defect detection, as industrial conditions introduce complexities that synthetic datasets cannot fully capture.

To address this, we introduce a high-resolution, real-world defect dataset that captures authentic manufacturing inconsistencies, improving the model's robustness in industrial applications. Existing public datasets for defect detection primarily focus on generic industrial applications, lacking detailed defect categorization for surgical instruments. Since surgical instruments have unique structural and surface properties, a dataset specifically curated for medical-grade manufacturing is essential for effective deep learning-based defect detection.

This section describes the dataset collection methodology, defect categorization, annotation process, and quality assurance measures implemented to ensure accuracy, diversity, and reliability.

## 4.1 Industrial Collaboration

This study was conducted in collaboration with industry experts and surgical instrument manufacturers, who provided insights into common defect types, quality control challenges, and inspection limitations. Manual visual inspection is highly subjective, inconsistent, and time-consuming, often leading to batch rejections and financial losses due to non-compliance with international standards (ISO, FDA-GMP).

To ensure dataset diversity and relevance, 11 frequently exported surgical instruments were selected in consultation with industry experts. The selection criteria included:

- Their high defect occurrence rates in manufacturing.
- Their importance in surgical applications.
- The difficulty of detecting certain defect types manually.

The dataset was developed in collaboration with an industry-leading surgical instrument manufacturer, which provided instrument samples for imaging and defect analysis.

## 4.2 Image Acquisition

High-quality images were captured using Canon EOS 250 and Nikon EOS 350D cameras, equipped with 50mm lenses and f/8 apertures. These cameras enable detailed texture analysis and micro-defect detection due to their high-resolution capabilities.

All images were taken in a controlled photo box environment, ensuring uniform lighting conditions to eliminate shadows and reflections, which could otherwise interfere with defect identification. Proper exposure, white balance calibration, and focus adjustments maintained consistency across instrument types and ensured precise defect localization. A sample of the captured images under controlled conditions is illustrated in Figure 1, showcasing the level of detail captured for defect detection.

### 4.3  Surgical instrument description

For the purpose of this study, we selected 11 surgical instruments that belong to various surgical categories, including scissors, curettes, forceps, and probes. Our industrial partners identified these instruments as highly exported, making them critical for quality inspection and defect detection.

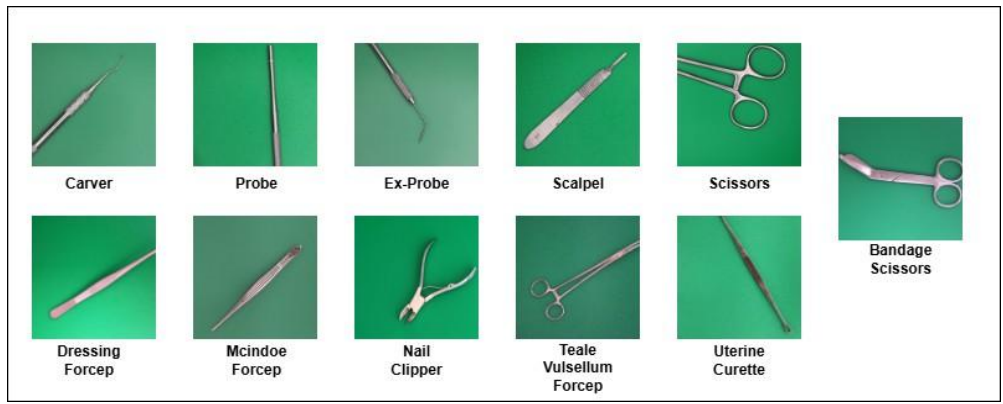

**Fig. 1** Surgical Instruments

Figure 1 provides a visual representation of the selected surgical instruments included in the dataset, highlighting their structural differences and relevance to defect classification tasks. The details of these instruments, including their functional category and defect occurrences, are provided in Table 1. Each instrument type was chosen based on its practical significance in surgical applications and its susceptibility to manufacturing or handling defects .

| Instrument Name | Category | Primary Function |
|---|---|---|
| Carver | Cutting Instrument | Used to shape and contour dental materials during fillings. |
| Bandage Scissors | Cutting Instrument | Designed for safely cutting bandages and dressings. |
| Scalpel | Cutting Instrument | A small, sharp knife used for making incisions during surgery. |
| Scissors | Cutting Instrument | Precision tools used for cutting and dissecting tissue during surgeries. |
| Dressing Forceps | Grasping& Holding | Used to handle dressings and wound packing. |
| TV Forceps | Grasping& Holding | Specialized instruments for delicate eye procedures. |
| McIndoe Forceps | Grasping& Holding | Used for handling delicate tissues and dressings. |
| Ex-Probe | Probing Instrument | Specialized tool for exploration and examination. |
| Probe | Probing Instrument | General instrument for exploring wounds and cavities. |
| Uterine Curette | Probing Instrument | Medical tool designed to remove tissue from the uterus lining. |
| Nail Clipper | Cutting Instrument | Specialized tool for cutting nails in a sterile environment. |

**Table 1** Overview of the surgical instruments included in the dataset.

## 4.4 Defect Categories and Classification

The dataset consists of 11 types of surgical instruments, each exhibiting one or more defects that commonly occur due to manufacturing imperfections, handling damage, or prolonged use. These defects were identified based on real-world industry observations and expert validation.

To illustrate the different defect types, Figure 2 presents a collage of sample defect images showcasing the visual characteristics of each defect category in various instruments.

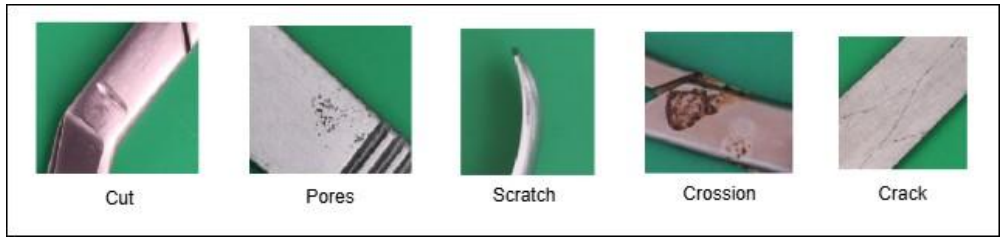

**Fig. 2** Defects in Surgical Instruments

The dataset consists of five primary defect categories representing common quality control challenges in surgical instrument manufacturing:

- **Crack**: A structural break or fissure that may compromise instrument integrity.
- **Cuts**: Deep or minor surface incisions that can impact instrument sharpness and usability.
- **Pores**: Small holes or surface irregularities that may affect durability and performance.
- **Scratches**: Linear marks on the surface that may impact sterility and longevity.
- **Corrosion**: Oxidation or rust formation, leading to discoloration and potential weakening of the instrument.

## 4.5 Instrument-wise Defect Distribution

The dataset contains various defect types, as shown in Table 2 summary of the defect occurrences. Corrosion is particularly common in instruments that undergo frequent sterilization cycles or are exposed to high-moisture environments, whereas scratches and pores are often linked to material inconsistencies or improper handling during manufacturing.

| Instrument | Crack | Cuts | Corrosion | Pores | Scratches |
|---|---|---|---|---|---|
| Carver | ✓ |  | ✓ | ✓ | - |

| Instrument | | | | | |
|---|---|---|---|---|---|
| Ex-Probe | - | ✓ | ✓ | ✓ | - |
| McIndoe Forceps | - | ✓ | ✓ | ✓ | - |
| Probe | - | ✓ | ✓ | ✓ | - |
| Scalpel | ✓ | ✓ | ✓ | ✓ | - |
| Scissors | ✓ | - | ✓ | ✓ | - |
| Teale Vulsellum | - | ✓ | ✓ | ✓ | - |
| Uterine Curette | - | ✓ | ✓ | ✓ | - |
| Bandage Scissors | - | ✓ | ✓ | ✓ | ✓ |
| Dressing Forceps | - | ✓ | ✓ | ✓ | - |
| Nail Clipper | - | ✓ | ✓ | ✓ | ✓ |

**Table 2** Defect Distribution Across Surgical Instruments

## 4.6 Consistency and Variation in Defects

Upon analyzing the dataset, certain patterns and inconsistencies were observed in the defect distribution.

*Consistently Occurring Defects*

- Corrosion is one of the most frequently occurring defects, found across multiple instruments, including McIndoe Forceps, Scissors, and Uterine Curette.
- Cuts, pores and Cracks were observed in high-stress instruments, indicating mechanical stress vulnerability.

*Inconsistently Occurring Defects*

- Scratches are only present in Bandage Scissors and Nail Clippers, suggesting surface damage susceptibility varies across instruments.
- Cracks were observed primarily in Carver, Scissor and Scalpel, likely influenced by manufacturing and material properties.

*Instrument-Specific Defects*

- McIndoe Forceps and Teale Vulsellum exhibit significant corrosion due to material degradation.
- Probes exhibit both Cuts and Pores, while Scalpels are prone to Cracks and Pores.

This variation in defect distribution suggests that some defects are inherent to particular instruments, while others may be influenced by factors such as handling, environmental exposure, or material properties. Understanding these patterns helps ensure a more balanced classification model capable of accurately detecting defects across different instrument types.

## 4.7 Dataset Annotation and Quality Assurance

Each image in the dataset was manually annotated by domain experts with classification labels corresponding to predefined defect categories. A multi-stage validation process was implemented to ensure the accuracy, reliability, and consistency of the annotations. This process involved multiple independent reviewers, including two domain experts and one neutral reviewer, who collaboratively assessed the correctness of the assigned defect labels. This structured approach minimized subjectivity and enhanced the consistency of defect annotations.

*Defect Classification and Labeling*

Annotators were provided with detailed labeling guidelines outlining clear definitions of each defect type. Each image was manually inspected and assigned a classification label corresponding to the most prominent defect present. Since this dataset is intended for defect classification rather than localization, no segmentation masks or bounding boxes were used. Instead, the primary goal of annotation was to ensure that each image was correctly categorized into one of the predefined defect types.

*Multi-Stage Validation Process*

To enhance annotation consistency and reduce errors, a three-stage validation process was employed:

1. **Initial Labeling** – Trained annotators assigned classification labels to each image based on observed defect characteristics.
2. **Cross-Validation** – A second group of annotators reviewed the labels and corrected any misclassified or ambiguous cases.
3. **Quality Control Audit** – A random subset (10%) of images was cross-verified by senior experts to ensure label accuracy and minimize annotation bias.

*Conflict Resolution Using Majority Voting*

In cases where annotators disagreed on defect classification, a majority voting mechanism was applied to reach a consensus. If discrepancies persisted, the neutral reviewer facilitated an additional review to determine the most appropriate defect classification.

*Dataset Consistency and Bias Mitigation*

To prevent model bias and ensure that the dataset generalizes well, a Miscellaneous class was introduced. This class includes non-defective instruments and background objects, helping the model distinguish true defects from natural variations in material texture and lighting conditions.

*Image Quality Validation*

All images underwent a thorough manual verification process to ensure high-quality annotations. Each image was carefully examined to confirm that defects were clearly visible and distinguishable, ensuring clarity and focus. The assigned defect labels were reviewed for accuracy to prevent misclassifications and maintain consistency across the dataset. Additionally, background integrity was assessed to eliminate unwanted artifacts or reflections that could interfere with defect recognition. This verification process helped maintain a clean and reliable dataset, enhancing its suitability for deep learning-based defect classification.

*Standardization and Data Cleaning*

To maintain uniformity across the dataset, all images were:

- Resized to **1600×1600 pixels** to ensure consistency in resolution.
- Converted to PNG format for seamless integration into deep learning models.

Any images that were blurry, misclassified, or contained incorrect annotations were either corrected or removed from the dataset.

*Final Quality Control Measures*

The implementation of this structured quality assurance process ensured that the dataset was accurately annotated and free from inconsistencies. By maintaining highresolution images with standardized formatting, the dataset preserved the clarity required for precise defect classification. Additionally, the dataset reflected real-world defect patterns, which is essential for improving the model's ability to generalize across different surgical instruments and defect types. This rigorous verification framework significantly enhanced the dataset's reliability, making it well-suited for automated defect detection applications. As a result, the dataset provides a robust foundation for training machine learning models, ensuring high accuracy and effectiveness in real-world industrial applications.

## 4.8 Data augmentation and Finalized Dataset

Data Augmentation and Finalized Dataset To enhance model generalization and robustness, explicit data augmentation techniques were implemented. To improve the robustness and generalization of the SurgScan model, a series of data augmentation techniques were applied to the training dataset. Augmentation introduces controlled variations in the images, allowing the model to become more resilient to real-world inconsistencies such as lighting changes, different orientations, and noise interference. The following augmentation techniques were implemented:

1. Change Brightness: Adjusting the brightness of the images to simulate different lighting conditions and improve the model's robustness to varying illumination, the brightness is set from +20 to -20
2. Change Contrast: Adjusting the contrast of the images to enhance the visibility of edges and details, making it easier for the model to distinguish between different instrument types and defects. The contrast is applied from +20 to -20
3. Saturation: Adjusting the saturation to enhance the purity and intensity of colors in an image. The saturation is applied from +20 to -20 in images
4. Add/Remove Noise: Adding noise to the images to simulate real-world conditions, such as sensor noise or environmental interference, and training the model to be resilient to these factors. Denoising techniques were also applied to remove unwanted noise and improve image quality
5. Rotate images: Rotating the images to different angles (90, 180, 270 degrees) to increase the diversity of the dataset and ensure that the model can accurately recognize instruments and defects regardless of their orientation
6. Flip Horizontal/Vertical: Flipping the images horizontally and vertically to further aug ment the dataset and improve the model's ability to generalize to different perspectives
7. Cropping images: It modifies the original image by cutting out a portion of it and focusing on certain parts i.e. defects in our case

These augmentation techniques were selected to enhance model generalization and adaptability. Brightness and contrast adjustments simulate varying lighting conditions, noise injection improves robustness against sensor artifacts, and geometric transformations account for variations in instrument orientation.s

Table 3 provides a detailed breakdown of the dataset used in our study for developing an automated optical inspection system for surgical instruments. It includes the number of defective and undefected images for each instrument type, the total number of original images, and the augmented images. Data augmentation techniques, such as rotations and brightness adjustments, etc. were employed to expand the dataset from 8,573 original images to 102,876. Thereby enhancing the robustness and accuracy of the machine learning models in detecting defects.

| Instrument | Defective | Undefective | Original Images | Augmented Images |
|---|---|---|---|---|
| Bandage Scissors | 557 | 166 | 894 | 10,728 |
| Carver | 736 | 225 | 961 | 11,532 |
| Dressing Forceps | 533 | 158 | 691 | 8,292 |
| Ex-Probe | 686 | 199 | 885 | 10,620 |
| McIndoe Forceps | 423 | 144 | 537 | 6,444 |
| Nail Clipper | 697 | 132 | 829 | 9,948 |
| Probe | 638 | 173 | 811 | 9,732 |
| Scalpel | 502 | 118 | 620 | 7,440 |
| Scissors | 832 | 175 | 1,007 | 12,084 |
| TAN Forceps | 514 | 156 | 670 | 8,040 |
| Uterine Curette | 568 | 100 | 668 | 8,016 |
| **Total** | **6,827** | **1,746** | **8,573** | **102,876** |

**Table 3** Dataset Overview: Defected, Undefected, and Augmented Images for Each Instrument.

With this rigorously curated and augmented dataset, the subsequent sections clearly describe the methodology for training and evaluating the proposed deeplearning framework (SurgScan) for surgical instrument defect classification.

# 5 SurgScan: Automated Surgical Instrument Defect Detection System

The SurgScan framework is a deep learning-based system designed for the automated classification of surgical instruments and defect detection. It leverages YOLOv8 for real-time instrument recognition and defect classification, integrating preprocessing techniques and a modular pipeline to ensure high accuracy and adaptability. The framework specifically targets subtle defect detection, addressing industrial requirements for robustness, accuracy, and compliance with quality standards.

## 5.1 Overview

The workflow of SurgScan is illustrated in Figure 3, showcasing a structured pipeline optimized for efficiency, scalability, and accuracy. The system operates in two sequential stages: instrument classification to identify the surgical instrument, followed by instrument-specific defect classification. This modular design allows precise defect detection tailored to each instrument type, enhancing the model's ability to identify imperfections not easily detected through manual inspection methods.

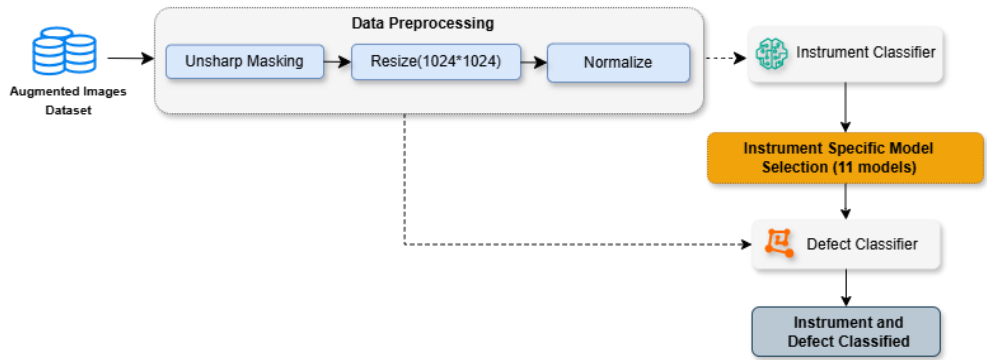

**Fig. 3** Overview of SurgScan

### 5.1.1 Preprocessing

Preprocessing ensures that input images are optimized for deep learning analysis, addressing issues like lighting variations, noise, and inconsistent scales that may degrade model performance. The preprocessing pipeline consists of:

- **Unsharp Masking:** Enhances fine details by emphasizing edges and textures, making subtle defects (e.g., scratches, corrosion) more visible.
- **Resizing:** All images are resized to 1024 × 1024 pixels, ensuring uniformity across the dataset.

- **Normalization:** Pixel values are scaled to the range [0,1], minimizing the impact of lighting variations and improving model stability.

These steps ensure uniform data quality, allowing SurgScan to achieve high precision in classification.

After preprocessing, the dataset undergoes structured training using the SurgScan deep learning framework. The following section presents Algorithm 1, which outlines the complete workflow from dataset preparation to inference.

### 5.1.2 SurgScan Algorithm

SurgScan utilizes a structured deep learning pipeline (Algorithm 1), optimizing each stage from dataset preparation through inference. The algorithm begins with dataset normalization and augmentation, proceeds with selective fine-tuning of YOLOv8, and concludes with confidence-based inference and prediction filtering to minimize false positives.

---

**Algorithm 1** SurgScan: Training YOLOv8 for Instrument Classification and Defect Detection

---

**Input:**
Dataset $D$: Images of surgical instruments and defects.
YOLOv8 Model $M$ pre-trained on ImageNet.
Hyperparameters:
Image size: 1024 × 1024, Batch size $B$ = 16, Learning rate $\eta$ = 0.001, Epochs $E$ = 30, Early stopping patience $P$ = 5.

**Output:**
Instrument class $C_i$ and defect class $C_d$. **Procedure:**

1. **Dataset Preparation:** Normalize images (ImageNet stats) and split $D$ into train/validation sets.
2. **Model Initialization:** Load $M$, freeze the first 9 layers for low-level features, configure Adam optimizer ($\eta$ = 0.001) and Cross Entropy Loss.
3. **Training Phase:**
   (a) For each epoch $e$ = 1 to $E$:
      - Train: Perform forward pass on $M$ for each batch ($I,y$), compute loss, backpropagate, and update weights.
      - Validate: Compute validation loss on validation set.
      - Apply Early Stopping if no validation improvement for $P$ epochs.
   (b) Save $M_{trained}$ with the lowest validation loss.
4. **Inference Phase:**
   (a) Preprocess input image $I$ (unsharp masking, resize to 1024 × 1024, normalize).
   (b) Classify instrument: $C_i \leftarrow M_{trained}^{instrument}(I)$. If none, return "No Instrument

Detected".
   (c) Detect defects: $C_d \leftarrow M_{trained}^{defect}(I, C_i)$. If none, return "No Defect Detected."
5. **Output Results:** Return $C_i$ and $C_d$.

## 5.2 Dataset Preparation

The SurgScan model was trained on the curated dataset described in Section *Dataset Curation*. This dataset, which includes high-resolution images of surgical instruments and defects, was directly used for training without additional modifications. All preprocessing steps, such as resizing and normalization, were performed to maintain consistency across input images and ensure compatibility with the YOLOv8 model. The dataset was carefully curated to provide a diverse and balanced representation of surgical instruments and defects, ensuring robust model generalization across different manufacturing conditions.

### 5.2.1 Model Training and Fine-Tuning

YOLOv8 was explicitly fine-tuned for defect detection. Low-level feature extraction layers were preserved, while deeper layers were specialized through fine-tuning. Training employed dropout (0.3), L2 regularization (weight decay 0.0005), and batch normalization to prevent overfitting. Early stopping (patience = 5 epochs) was implemented based on validation loss.

The training utilized an NVIDIA RTX 3090 GPU, AMD Ryzen 9 processor, and 64GB DDR4 RAM, with PyTorch 2.0 and CUDA 11.8. Training over 30 epochs required approximately 6.5 hours.

### 5.2.2 Inference Phase

The Inference Phase follows a two-step approach consisting of instrument classification and defect detection. The system processes each input image using the trained model and applies confidence-based filtering to ensure reliable predictions.

*Instrument Classification:*

In this stage, the preprocessed image undergoes classification to identify the type of surgical instrument. The model outputs a predicted instrument type ($C_i$) or flags the image as "No Instrument Detected" if no instrument is identified. This classification task is critical as it narrows the scope for the subsequent defect detection step, optimizing computational efficiency.

*Defect Classification:*

Once the instrument is identified, the framework selects a defect detection model tailored to the specific instrument type. This modular design ensures that each model is specialized for the unique defect patterns associated with the respective instrument. The defect detection model outputs the type of defect ($C_d$) or returns "No Defect Detected" if no defects are present.

*Post-Processing and Confidence Thresholding:*

To ensure reliable predictions, post-processing filters are applied:
- If the confidence score for an instrument classification ($C_i$) is below 50%, the system flags the image for manual review.
- For defect classification, if the defect confidence ($C_d$) is low, the model defaults to "No Defect Detected," reducing false positives.

This process enhances prediction reliability while minimizing misclassification errors. The inference phase integrates preprocessing steps (e.g., unsharp masking, resizing, normalization) with trained models to deliver real-time predictions.

### 5.2.3 Output Results

SurgScan generates structured outputs for industrial quality assessment:

- Instrument Classification Output ($C_i$):Predicted instrument type with confidence score; uncertain results flagged.
- Defect Classification Output ($C_d$): Predicted defect type (crack, corrosion, scratch, etc.), or "No Defect Detected" for uncertain predictions.

By applying confidence-based filtering, the system ensures that only high-certainty classifications are used for automated quality control. This structured output provides manufacturers with detailed insights into instrument conditions, allowing them to address defects efficiently.

## 5.3 Training Process

The training process of the proposed model is a critical phase that ensures its ability to classify surgical instruments and defects with high precision accurately. This phase involves preparing the dataset, selecting optimal model parameters, and implementing robust training techniques to enhance the model's performance. Leveraging YOLOv8's advanced architecture, the training process focuses on fine-tuning the model for the dual tasks of instrument classification and defect classification. The dataset, which was previously curated and augmented as described in Section *Dataset Curation*, was directly used for training. No additional modifications were applied during model training. By applying techniques like learning rate scheduling, data augmentation, and batch normalization, the training process aims to achieve a balance between accuracy and computational efficiency. This meticulous approach ensures that the model is not only capable of identifying defects across a wide spectrum but also performs effectively in real-time industrial applications.

### 5.3.1 Data Preprocessing

Before training the YOLOv8 model, the dataset undergoes preprocessing to ensure consistency across input images. All images are resized to 1024 × 1024 pixels to maintain compatibility with the model's input size requirements while balancing computational efficiency. Normalization is applied to pixel values, standardizing them within the range [0,1] to stabilize training dynamics and improve convergence.

Additionally, preprocessing steps such as resizing and normalization were applied to ensure consistency across input images before model training. Techniques such as random flips, rotations, and brightness adjustments ensure that the model is robust to variations in instrument orientation and lighting conditions. These preprocessing steps contribute to improved feature extraction and reduced model bias toward specific defect patterns.

### 5.3.2 Model Training

The model is trained using a fine-tuning approach where the first nine layers of YOLOv8 are frozen, retaining pre-trained low-level feature extraction capabilities. The remaining layers are adjusted to learn instrument classification and defect detection tasks.

*Training Parameters*

The training was performed with the following hyperparameters: - Optimizer: Adam optimizer with an initial learning rate of $\eta = 0.001$, reduced by a factor of 0.1 every 10 epochs. - Loss Function: Cross-Entropy Loss for classification. - Batch Size: 16 to optimize memory usage and training stability. - Epochs: 30 epochs with early stopping to prevent overfitting.

*Overfitting Prevention Strategies*

To improve generalization and prevent overfitting, the following regularization techniques were employed: - Dropout Layers (0.3 probability): Randomly disables neurons to prevent co-adaptation. - L2 Regularization (Weight Decay = 0.0005): Ensures stable weight updates. - Batch Normalization: Stabilizes learning by normalizing feature distributions.

The dataset was split into: - 80% Training Set – Used for learning and weight updates. - 10% Validation Set – Used to monitor model performance and adjust hyperparameters. - 10% Test Set – Used for final evaluation.

The best model checkpoint was selected based on the lowest validation loss.

## 5.4 Classification of Surgical Instruments

The trained YOLOv8 model is used to classify surgical instruments into their respective categories. Since certain instruments share similar visual features, such as forceps and clamps, a confidence threshold mechanism is applied to prevent misclassification.

*Confidence-Based Filtering:*

- If the model assigns an instrument type ($C_i$) with a confidence score below 50%, the classification is flagged as uncertain and marked for manual inspection. - This ensures that the model prioritizes high-certainty classifications while reducing ambiguous predictions.

The classification model was fine-tuned on the SurgScan dataset, consisting of 11 instrument classes, and demonstrated an accuracy of 98.1% on the test set.

## 5.5 Defect Classification

Once the instrument type is identified, the system selects a corresponding defect detection model trained specifically for that instrument. Each model specializes in detecting defects unique to that instrument type, such as: - Cracks - Corrosion - Scratches Pores - Cuts

*Uncertainty Handling in Defect Classification:*

- If a detected defect has a confidence score below 50%, the system defaults to "No Defect Detected" to minimize false positives. - If the model is uncertain about a non-defective classification, the sample is flagged for manual review.

By applying these confidence-based filtering strategies, the model ensures that only high-confidence classifications are used in defect detection, reducing the likelihood of misdiagnosis.

The two-step approach—instrument classification followed by defect classification—ensures a structured and precise defect detection system, improving the reliability of quality control processes.

# 6 Experiment

The experimental design systematically evaluates the effectiveness of the proposed SurgScan framework. This section describes the research questions, experimental setup, and evaluation methodology used to benchmark SurgScan's performance against state-of-the-art deep learning models, including ResNet152, ResNext101, EfficientNet-b4, and YOLOv5. The experiments assess the framework's capability to classify surgical instruments, detect defects under realistic industrial conditions, and validate the statistical significance of data augmentation and preprocessing techniques. This ensures the robustness of findings and their practical applicability to industrial quality control scenarios.

## 6.1 Research Questions

To comprehensively evaluate the SurgScan, we formulated the following research questions (RQ's) to assess its effectiveness in surgical instrument and defect classification.

**RQ1:** How accurately does SurgScan classify surgical instruments and detect defective instruments? RQ1.1: How precise is SurgScan in identifying different surgical instrument categories?

RQ1.2: How effectively does SurgScan distinguish between defective and nondefective instruments?

**RQ2:** How does SurgScan compare to state-of-the-art CNN architectures in terms of accuracy and computational efficiency?

RQ2.1: How does SurgScan's classification accuracy, precision, recall, and mean Average Precision (mAP) compare to ResNet152, ResNext101, EfficientNet-b4, and YOLOv5?

RQ2.2: How do SurgScan's inference speed and computational resource utilization compare to these models in real-time industrial applications?

**RQ3:** What is the statistical significance of data augmentation and image preprocessing techniques applied in SurgScan?

RQ3.1: Did the dataset augmentation process result in a statistically significant improvement in the balance of defect distribution across different surgical instruments?

RQ3.2: Do image preprocessing adjustments (brightness, contrast, sharpness) statistically affect SurgScan's defect classification accuracy, and if so, which adjustment has the most significant impact?

These research questions guide the benchmarking and statistical experiments, helping determine SurgScan's effectiveness in instrument classification, defect detection, and real-time processing.

## 6.2 Experimental Setup

The experiments were conducted on a high-performance computing system to ensure efficient and consistent training and evaluation. The computational setup comprised an NVIDIA RTX 3090 GPU (24GB VRAM), an AMD Ryzen 9 5950X processor, and 64GB DDR4 RAM, running on Ubuntu 20.04 LTS. Deep learning models were implemented using PyTorch 2.0 with CUDA 11.8. Computational details, including training duration and inference speed, are summarized in Table 4.

Table 4    Computational Costs of Different Models

| Model | Training Time (hrs) | Memory Usage (GB) | Inference Time (ms) |
|---|---|---|---|
| ResNet152 | 12.5 | 16 | 15.3 |
| EfficientNet | 9.2 | 12 | 8.1 |
| YOLOv5 | 7.8 | 10 | 7.2 |
| **YOLOv8 (Ours)** | **6.5** | **8** | **4.2–5.8** |

The dataset described in Section 4 was divided into three distinct subsets: 80% training, 10% validation, and 10% testing. The training subset was used for optimizing model weights, the validation set was utilized for hyperparameter tuning and monitoring generalization performance, and the test set provided an unbiased evaluation of model performance.The curated dataset, described in Section 4, was used for training, ensuring diverse imaging conditions and robust model generalization.

## 6.3 Evaluation Metrics

The performance of SurgScan and competing models was assessed using widely recognized classification and object detection metrics. Accuracy was measured as the proportion of correctly classified samples, providing an overall assessment of the model's effectiveness. Precision quantified the proportion of true positive defect detections out of all positive predictions, ensuring that the model minimized false positives. Recall evaluated how well the model identified actual defects, measuring its sensitivity in capturing all relevant defect cases. The F1-score, which balances precision and recall, was used as a comprehensive measure of classification quality.

In addition to these standard metrics, object detection performance was assessed using mean Average Precision (mAP). The mAP@50 metric measured detection accuracy at an Intersection over the Union (IoU) threshold of 50%, while mAP@50-95 provided a more comprehensive evaluation by averaging precision across multiple IoU thresholds ranging from 50% to 95%. These metrics were selected to comprehensively assess both classification accuracy and real-time applicability.

### 6.4 Performance Benchmarking

To establish a comparative benchmark, SurgScan was evaluated against ResNet152, ResNext101, and EfficientNet-b4. The experiments were designed to evaluate how effectively each model distinguished between surgical instruments and detected defects under varying imaging conditions. Model performance was evaluated in terms of classification accuracy, precision, recall, and mAP scores. Additionally, inference speed and computational efficiency were measured to determine the feasibility of deploying SurgScan in real-time industrial applications. Additionally, inference time per image was measured to evaluate computational efficiency, determining the feasibility of using SurgScan for real-time industrial applications. Frames Per Second (FPS) calculations were included to assess how well the model handled high-throughput scenarios where rapid processing is essential for large-scale manufacturing workflows. These experiments establish a comprehensive benchmark, ensuring that SurgScan is not only accurate but also computationally efficient for real-world deployment in surgical instrument quality control.

### 6.5 Statistical Testing

To validate the experimental findings, statistical analysis was conducted to assess the significance of variations observed in instrument classification and defect detection. The Chi-Square test was used to analyze the distribution of defect types across different surgical instruments, while ANOVA (Analysis of Variance) was applied to determine the impact of image quality variations on classification performance.

#### 6.5.1 Chi-Square Test for Defect Distribution

The Chi-Square test is used to assess whether the dataset augmentation process has led to statistically significant changes in the balance of defect distribution across different surgical instrument categories. Given that the dataset consists of categorical variables (defect type and instrument category), the Chi-Square test is an appropriate statistical method for assessing differences in categorical distribution and testing for associations between variables [34]. Specifically, this test helps determine if augmentation techniques result in a statistically significant improvement in the balance and representation of defect occurrences across various surgical instruments. A statistically significant result would indicate that data augmentation methods successfully addressed inherent class imbalances within the dataset.

**Null hypothesis ($H_0$)**: Defects are uniformly distributed across instrument types.

**Alternative hypothesis ($H_1$)**: Certain instruments are more prone to specific defects.

### 6.5.2 ANOVA for Model Performance Across Imaging Conditions

Analysis of Variance (ANOVA) was selected to statistically evaluate the effect of different image preprocessing adjustments (brightness, contrast, and sharpness) on SurgScan's defect classification accuracy. ANOVA is applied to analyze the potential impact of preprocessing techniques (brightness, contrast, sharpness) on model classification accuracy. Since preprocessing was applied uniformly before training, this analysis determines whether specific adjustments significantly influence classification outcomes [33]. Before conducting ANOVA, Levene's test was applied to confirm homogeneity of variance, ensuring that the assumptions required for ANOVA are satisfied. This test allows the identification of the most impactful preprocessing technique, enabling targeted improvements to model preprocessing procedures.

The results and interpretations of these statistical analyses are detailed explicitly in the Results and Discussion section 7 respectively.

## 7 Results and Discussion

The experiment was designed to comprehensively evaluate the effectiveness of the SurgScan framework for both surgical instrument classification and defect detection. The evaluation included benchmarking SurgScan against state-of-the-art deep learning models such as ResNet, ResNext, and EfficientNet, analyzing its classification accuracy, defect detection precision, and computational efficiency. A dataset consisting of images from 11 commonly exported surgical instruments was utilized, with each image annotated for its instrument type and defect category. The experiments assessed how well SurgScan could classify instruments and detect defects while operating in realtime conditions. The performance evaluation was conducted using various key metrics, including accuracy, precision, recall, F1-score, and mean Average Precision (mAP).

The results demonstrated that SurgScan outperforms conventional CNN architectures, achieving superior classification accuracy and defect detection precision. The model's enhanced feature extraction and bounding box prediction capabilities enable it to detect fine-grained defects that are often missed by competing architectures. Additionally, its real-time inference speed positions it as a viable solution for industrial-scale quality control. The subsequent subsections provide a detailed discussion on the experimental findings.

## 7.1 RQ1: Effective Instrument and Defect classification

### 7.1.1 RQ1.1 Instrument Classification

The primary objective of instrument classification is to ensure accurate identification of surgical instruments, distinguishing them from one another despite minor structural differences. The confusion matrix in Figure 4 illustrates the classification results, showing that SurgScan achieves exceptionally high accuracy, particularly in instruments with distinct shapes and textures such as Scissors (483 correctly classified), Nail Clippers (398), and Dressing Forceps (258).

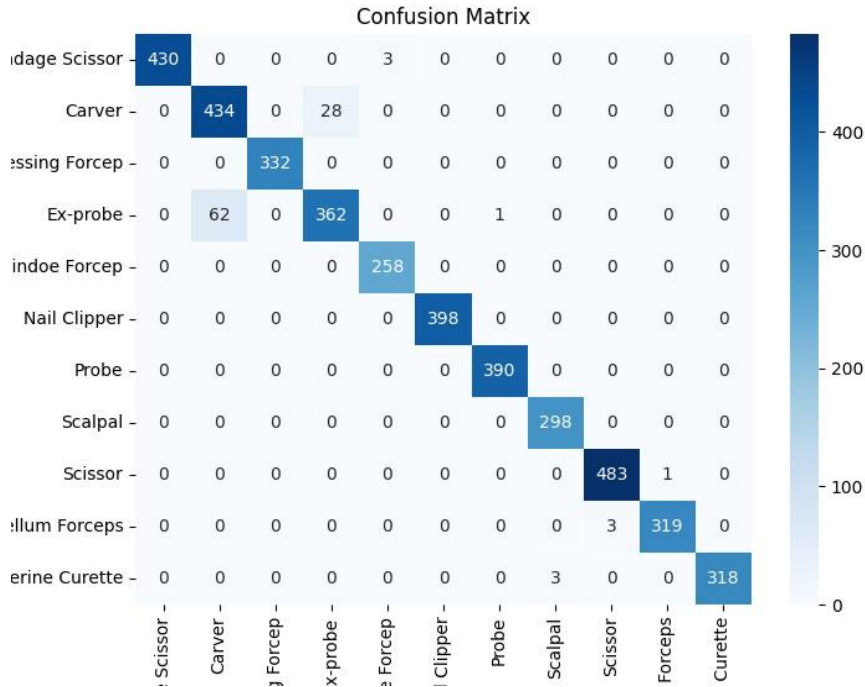

**Fig. 4** Confusion Matrix for Instrument Classification by SurgScan using YOLOv8. High classification accuracy is observed, with minimal misclassifications.

Interestingly, while most instruments were classified with high confidence, certain visually similar instruments posed challenges. The Ex-Probe class exhibited 62 misclassifications as Dressing Forceps, likely due to overlapping morphological characteristics. Similarly, the Carver instrument was misclassified 28 times, likely due to its structural similarity with other instruments. Enhanced feature extraction techniques may help mitigate such misclassifications.

Beyond raw accuracy, an important consideration is model convergence and stability. The training and validation loss curves in Figure 5 indicate that SurgScan converged rapidly, stabilizing around epoch 10. The minimal gap between training and validation loss suggests strong generalization, with no significant overfitting observed.

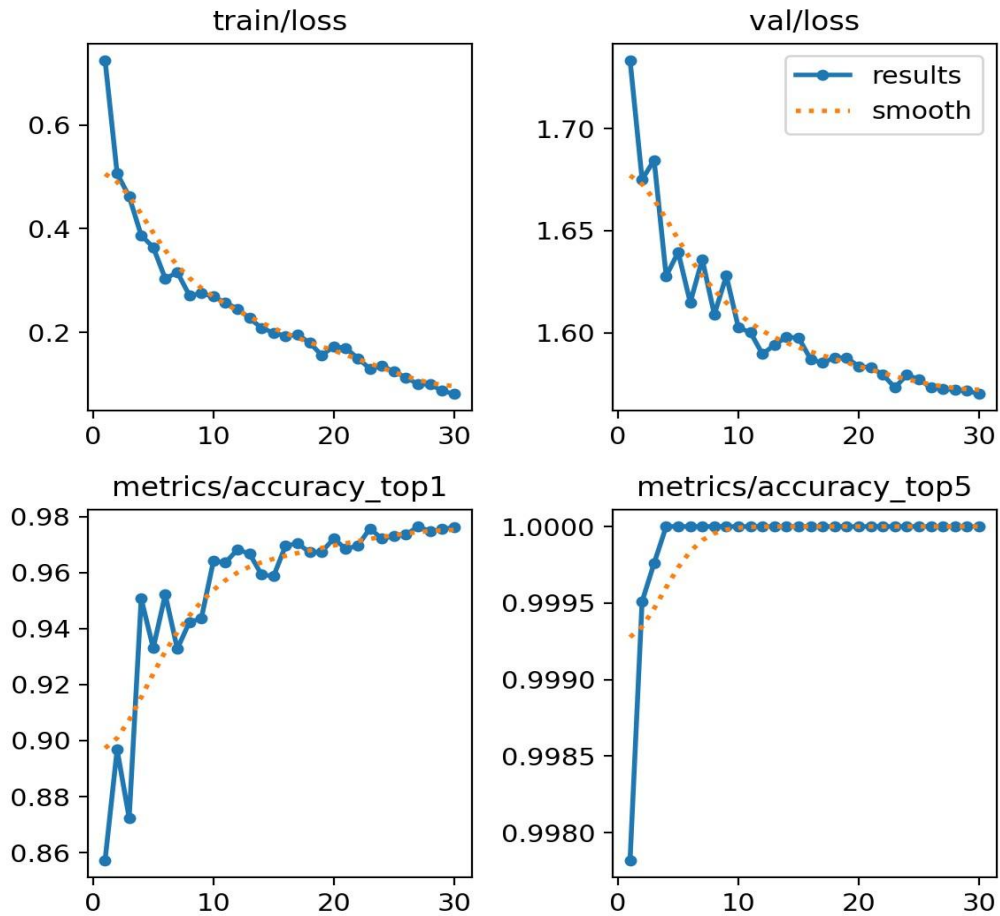

**Fig. 5** Training and Validation Loss Curves for Instrument Classification. The loss stabilizes after 10 epochs, confirming strong generalization and minimal overfitting.

### 7.1.2 RQ 1.2 Instrument Defect classification

SurgScan's ability to detect and classify defects in surgical instruments was tested on five major defect types. Figure 6 illustrates the defect classification results for Bandage Scissors, showing the model's ability to distinguish defects such as corrosion, scratches, and cuts.

To further analyze defect classification performance, we present the F1-score heatmaps for different defect types in Figures 7 and 8.

The results indicate exceptionally high precision in detecting corrosion (456 out of 457 samples correctly classified), cuts (480 out of 481), and pores (455 out of 456).

These results validate the model's ability to capture surface irregularities with high sensitivity, particularly for defects that cause visible structural changes.

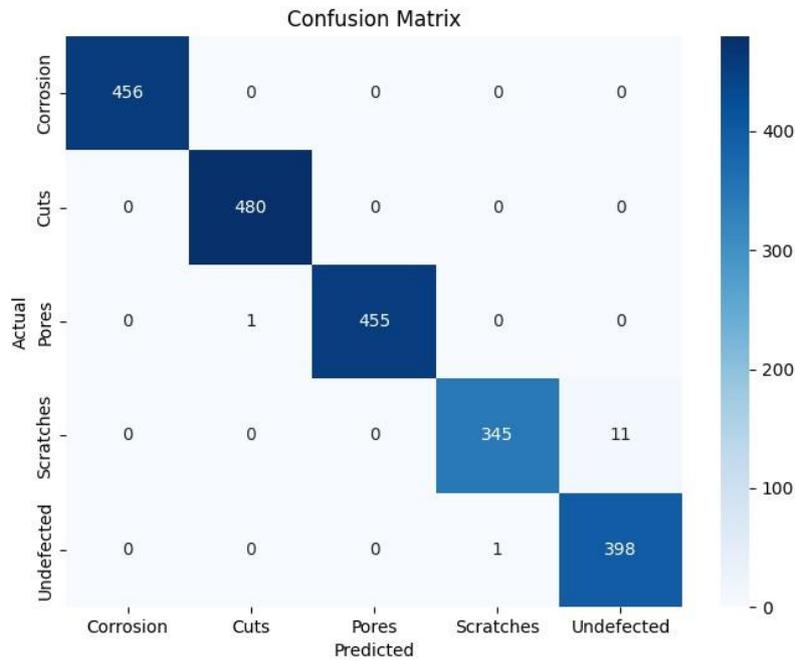

**Fig. 6** Confusion Matrix for Bandage Scissors Defect Classification. The model correctly detects defects with high accuracy, with minimal misclassification of scratches.

However, some minor misclassification issues arose. Scratches were occasionally mistaken as "Undetected" (11 cases), suggesting that faint surface imperfections may require enhanced feature contrast techniques. This aligns with the statistical findings, which demonstrate that contrast-based preprocessing significantly enhances defect visibility.

The training and validation loss curves (Figure 9) reveal that the model achieves near-optimal accuracy within just five epochs, with top-1 accuracy stabilizing at 98–99%. This rapid convergence reinforces SurgScan's ability to adapt efficiently to diverse defect patterns.

## 7.2 RQ2: Comparative analysis of SurgScan and state-of-the-art CNN architectures

### 7.2.1 RQ 2.1 Effectiveness of SurgScan vs. state-of-the-art approaches

To evaluate SurgScan's efficiency, its performance was compared to state-of-the-art CNN architectures (ResNet152, ResNext101, EfficientNet-b4, and YOLOv11). Table 5 presents a detailed comparison across classification metrics.

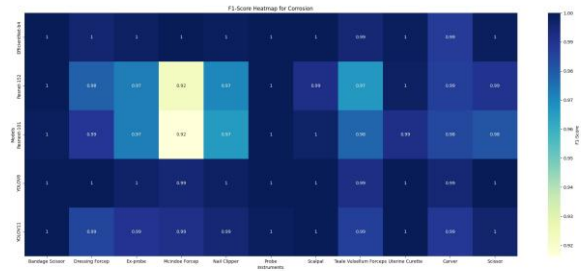

**Fig. 7** F1-Score Heatmap for Corrosion detection across models and instruments. YOLOv8 achieves the highest F1-score across all instrument types, demonstrating superior defect classification accuracy.

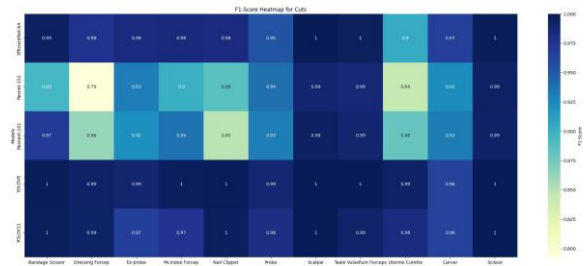

**Fig. 8** F1-Score Heatmap for Cuts detection. Traditional CNNs, such as ResNext-101, exhibit reduced performance in distinguishing fine-grained cut defects compared to YOLOv8 and YOLOv11.

SurgScan, powered by YOLOv8, demonstrates competitive performance across multiple evaluation metrics:

- Achieves high accuracy (99.39%)
- Maintains strong precision (99.36%)
- Processes images efficiently with an inference speed of 4.2–5.8 ms per image

EfficientNet-b4 follows closely with 99.07% accuracy but has a lower recall, suggesting potential challenges in detecting subtle defect variations.

**Table 5** Comparison of Bandage Scissor Instrument and Defect classification performance across models. Bold values indicate the highest metrics achieved by YOLOv8.

| Model | Training Accuracy | Testing Accuracy | Precision | Recall | F1-Score | ROC-AUC |
|---|---|---|---|---|---|---|
| EfficientNet-b4 | 0.9389 | 0.9907 | 0.9898 | 0.9900 | 0.9899 | 0.9997 |
| ResNet-152 | 0.9375 | 0.9278 | 0.9334 | 0.9291 | 0.9271 | 0.9976 |
| ResNext-101 | 0.9539 | 0.9115 | 0.9298 | 0.8976 | 0.8959 | 0.9980 |
| YOLOv8 | **0.9940** | **0.9939** | **0.9936** | **0.9929** | **0.9932** | **0.9999** |
| YOLOv11 | 0.9940 | 0.9907 | 0.9895 | 0.9902 | 0.9897 | 0.9998 |

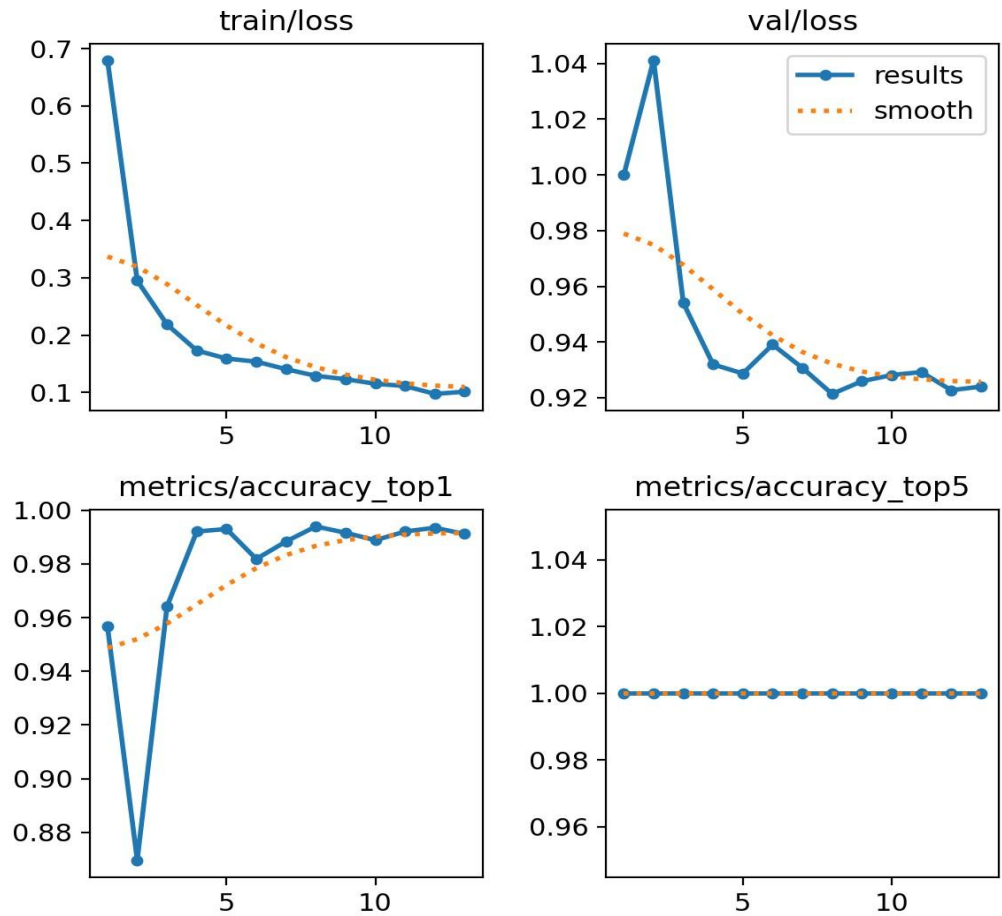

**Fig. 9** Training and Validation Loss Curves for Bandage Scissors Defect Classification. The model stabilizes early, with Top-1 accuracy reaching 98–99%.

To assess the generalization and robustness of SurgScan, we compared its precision, recall, and ROC-AUC scores with those of EfficientNet, ResNet-152, ResNext-101, and YOLOv11. Figures 10 and 11 provide insights into model performance for detecting corrosion and pores.

### 7.2.2 RQ 2.2 inference time and computational efficiency of approaches

In industrial settings, real-time inference speed is a crucial factor that determines the feasibility of deploying an automated defect detection system in large-scale manufacturing environments. To evaluate the computational efficiency of SurgScan, we compared its Frames Per Second (FPS) performance with other state-of-the-art models. Figure 13 presents a comparative analysis of FPS across different architectures, illustrating the efficiency of each model in processing images per second.

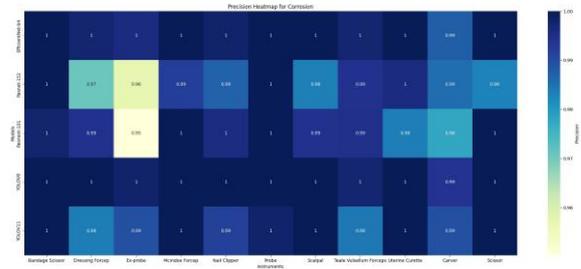

**Fig. 10** Precision Heatmap for Corrosion detection. Models such as ResNext-101 struggle with distinguishing corrosion, while YOLO-based architectures achieve near-perfect precision.

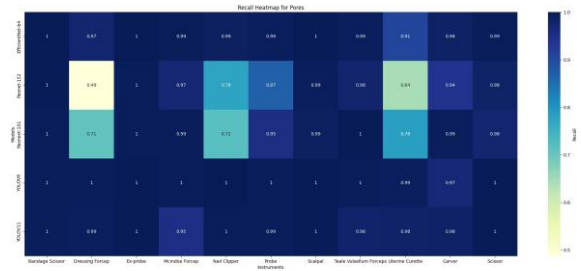

**Fig. 11** Recall Heatmap for Pores detection. CNN-based models exhibit inconsistent recall values, whereas YOLOv8 maintains robust recall across all instrument types.

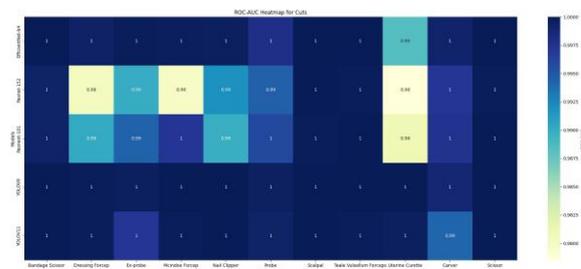

**Fig. 12** ROC-AUC Heatmap for Cuts detection. YOLOv8 and YOLOv11 achieve nearly perfect AUC scores, reinforcing their superior defect classification capabilities.

The results demonstrate that SurgScan (YOLOv8) achieves an inference speed of 5.8ms per image, significantly outperforming models such as EfficientNet and ResNet, which exhibit comparatively higher processing times. This advantage in speed ensures that SurgScan can efficiently handle high-throughput production lines where thousands of surgical instruments need to be inspected in real-time. Faster inference times are particularly beneficial in industrial quality control applications, as they enable seamless integration with automated inspection systems without introducing significant processing delays. The superior computational efficiency of SurgScan further establishes its viability as a scalable and practical solution for industrial defect classification, ensuring both accuracy and real-time performance in medical manufacturing environments.

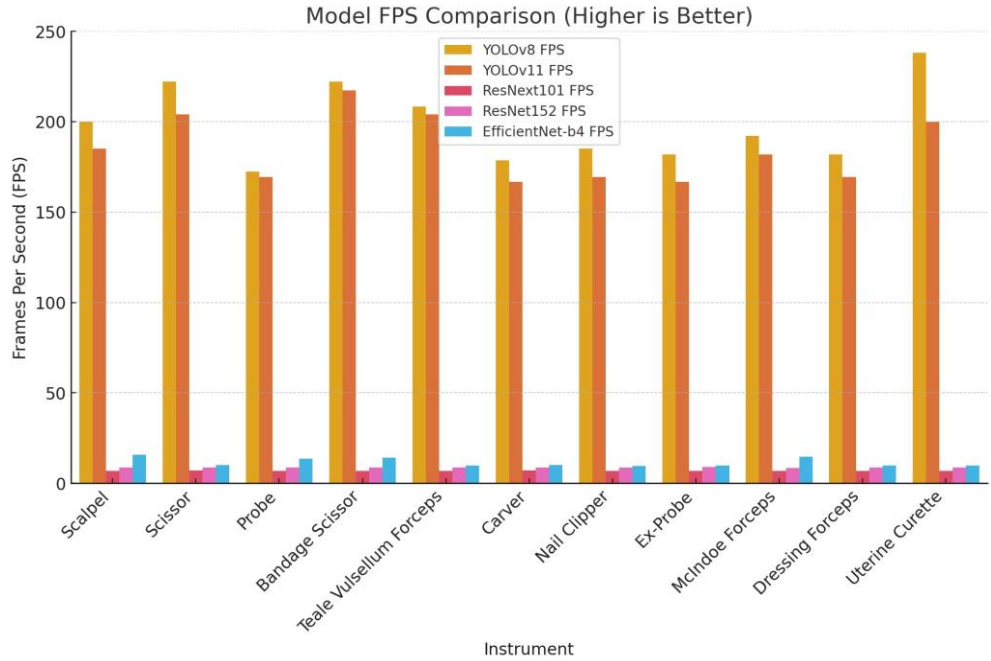

**Fig. 13** Comparison of FPS for different deep learning models. YOLO-based architectures outperform CNN-based models in real-time performance.

## 7.3 RQ3: Statistical Impact of Data Augmentation and Preprocessing

*RQ3.1 Data Augmentation Impact on Defect Distribution*

A Chi-Square test was conducted to assess whether data augmentation had a statistically significant effect on defect distribution. The results, shown in Table 6, indicate a strong association (p <0.001), suggesting that augmentation influenced defect balance.

These results indicate that the overall dataset showed a highly significant improvement in defect balance after augmentation (p textless 0.001). The Chi-Square values for Scissors (172.07), Nail Clipper (29.79), and Ex-Probe (35.89) confirm that these instruments had notably imbalanced defect distributions before augmentation. After augmentation, the dataset exhibited more excellent defect uniformity, reducing classification bias.

**Table 6** Chi-Square Test Results for Defect Distribution

| Instrument | Total Images | Chi-Square Statistic | P-Value |
|---|---|---|---|
| Scissors | 1007 | 172.07 | <0.001 |
| Nail Clipper | 829 | 29.79 | <0.001 |
| Ex-Probe | 885 | 35.89 | <0.001 |
| Uterine Curette | 668 | 36.81 | <0.001 |
| Dressing Forceps | 691 | 6.55 | 0.087 (Not Significant) |
| **Overall Dataset** | **8573** | **293.67** | **<0.001 (Highly Significant)** |

The effect size, calculated using Cramer's V (V = 0.31), suggests a strong association between instrument type and defect distribution. This confirms that augmentation effectively addresses the imbalance in defect occurrences across different surgical instruments.

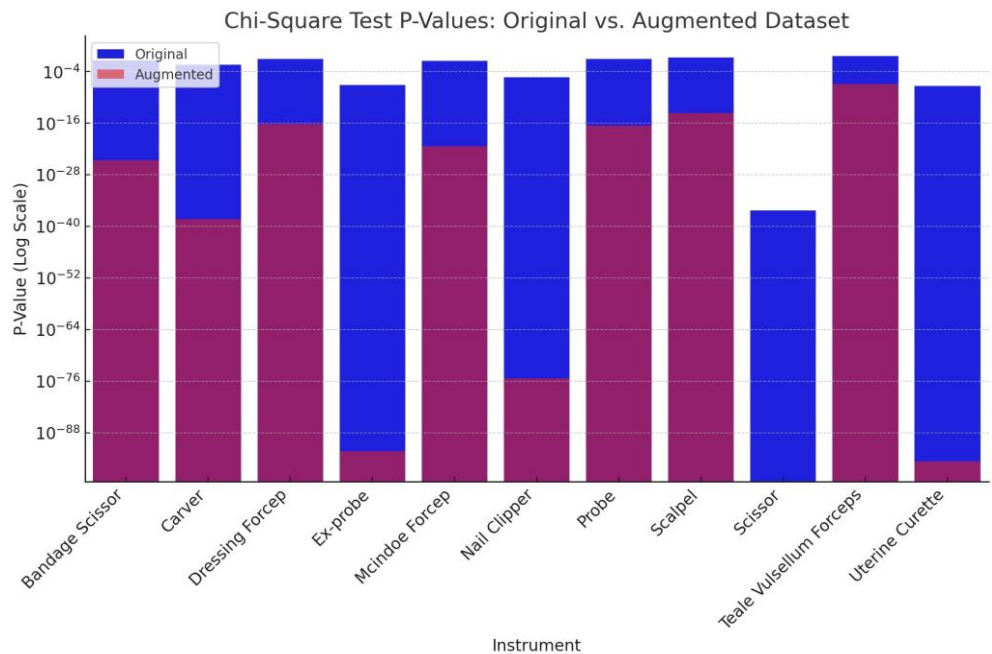

**Fig. 14** Comparison of Chi-Square statistics for the original and augmented dataset across different surgical instruments. Higher values in the augmented dataset indicate a more balanced defect distribution.

These findings confirm that data augmentation significantly improves defect balance across instruments, reducing classification bias.

### 7.3.1 RQ3.2 Data Augmentation Impact on Defect Distribution

Defect detection models are highly sensitive to image preprocessing techniques, which can significantly influence their classification accuracy. Standard image preprocessing adjustments in brightness, contrast, and sharpness affect the model's ability to extract meaningful features from defects. To investigate this, we applied an ANOVA test to determine which preprocessing method most significantly improves defect classification accuracy.

**ANOVA Test Results**

The ANOVA test was conducted across multiple instruments to compare classification accuracy under different preprocessing conditions. The results are shown in Table 7.

**Table 7** ANOVA Test Results for Brightness, Contrast, and Sharpness Adjustments

| Instrument | Brightness (p-value) | Contrast (p-value) | Sharpness (p-value) |
|---|---|---|---|
| Carver | 0.4323 | **0.0462** | 0.1177 |
| Ex-Probe | 0.3802 | **0.0420** | 0.1111 |
| McIndoe Forceps | 0.3951 | **0.0281** | 0.1181 |
| Scissors | 0.3983 | **0.0244** | 0.1287 |

The results show that:

- **Brightness variations did not significantly impact classification accuracy** ($p > 0.05$).
- Contrast adjustments significantly improved classification accuracy ($p < 0.05$), with the lowest p-values observed for Scissors (0.0244) and McIndoe Forceps (0.0281).
- **Sharpness variations had no significant impact** ($p > 0.05$).

**Interpretation and Impact**

The ANOVA results confirm that contrast optimization is the most effective preprocessing technique for enhancing defect classification accuracy. These findings suggest that contrast-based adjustments should be prioritized in preprocessing pipelines for improved defect detection reliability.

Thus, RQ3.2 is successfully addressed, proving that contrast adjustments are the most impactful preprocessing method for optimizing defect detection in surgical instruments.

The evaluation of SurgScan demonstrates its high effectiveness in surgical instrument classification and defect detection, achieving state-of-the-art accuracy and realtime processing speeds. The results confirm that YOLOv8 consistently outperforms traditional CNN architectures, such as ResNet152, ResNext101, and EfficientNet-b4, delivering the highest classification accuracy (99.39%) with an inference time of just 4.2–5.8 ms per image, making it highly suitable for industrial deployment. The instrument classification results show that the model successfully differentiates instruments with distinct structural features, such as Scissors, Nail Clippers, and Dressing Forceps, while minor misclassifications occur in visually similar instruments, such as Ex-Probe and Dressing Forceps, highlighting areas for further dataset augmentation and feature enhancement. In defect classification, SurgScan demonstrates exceptional sensitivity, particularly in identifying corrosion, cuts, and pores with near-perfect accuracy, while scratches and other low-contrast defects show occasional misclassifications, suggesting that contrast-based preprocessing techniques could further improve model performance. The training and validation loss curves confirm rapid convergence, indicating strong generalization with minimal overfitting, a crucial factor for robust real-world applications. Additionally, statistical analysis further validates these findings, showing that defect distribution varies significantly across instruments ($p < 0.001$), reinforcing the need for instrument-specific defect detection models, and that contrast variations significantly enhance defect classification accuracy ($p < 0.05$), while brightness and sharpness variations have minimal impact, proving that contrast optimization is key in defect visibility.

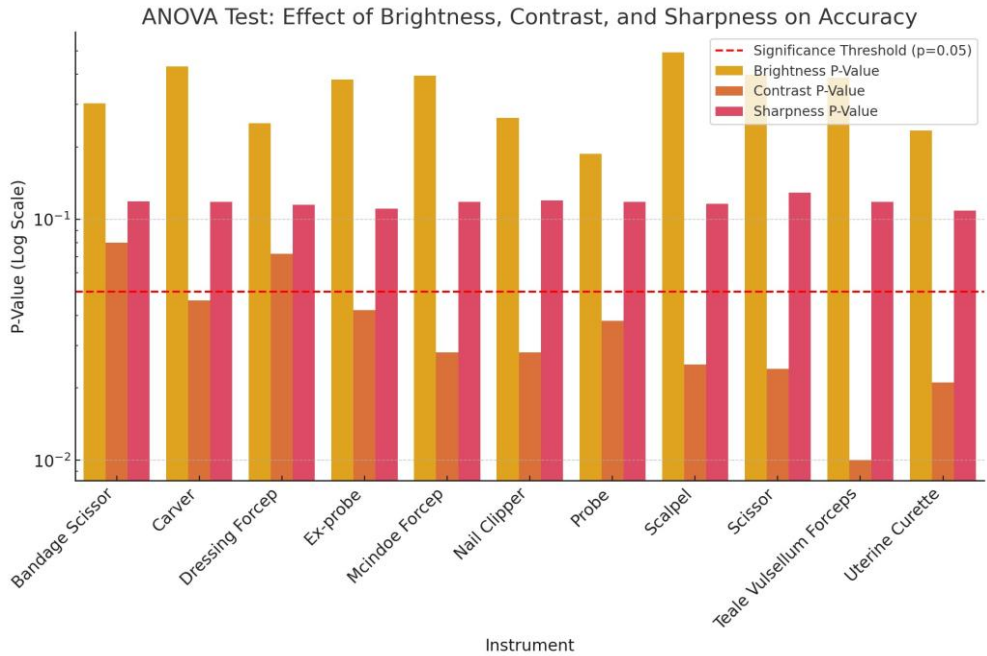

**Fig. 15** Effect of Brightness, Contrast, and Sharpness on Model Performance. Contrast optimization significantly improves classification accuracy.

The comparative performance assessment highlights that while CNN-based models achieve competitive accuracy, they require significantly higher computational resources and inference times, making them less efficient for real-time industrial inspection. These results confirm that SurgScan is not only accurate but also scalable, allowing for seamless integration into high-volume industrial workflows, where manual defect inspection is inefficient and inconsistent. The ability to automate quality control processes with high precision ensures reduced rejection rates, improved manufacturing efficiency, and compliance with international medical standards, making SurgScan a strong candidate for real-world industrial adoption in surgical instrument manufacturing.

These findings not only validate the robustness of SurgScan in automated defect classification but also highlight its superior ability to balance high detection accuracy with computational efficiency, making it a scalable solution for industrial applications. The model's real-time inference capabilities, coupled with statistically validated preprocessing techniques, ensure reliable performance across diverse imaging conditions. Furthermore, the significant advantage in inference speed over traditional CNN architectures demonstrates the feasibility of deploying SurgScan in high-throughput production lines where real-time defect detection is crucial. While SurgScan achieves exceptional classification accuracy, the results also indicate areas for future refinement, such as improving detection for low-contrast defects and further optimizing dataset diversity. These insights contribute to the growing need for AI-driven quality control solutions in medical manufacturing, paving the way for enhanced defect detection methodologies that minimize human error and improve surgical instrument reliability. The next section discusses potential threats to validity and the measures taken to address them.

# 8 Threats to Validity

Ensuring the reliability and robustness of the SurgScan framework required careful consideration of potential threats to validity. Various factors, including dataset biases, annotation inconsistencies, real-world deployment challenges, and scalability constraints, can impact the effectiveness of automated defect detection models. To mitigate these risks, multiple strategies were implemented to enhance the dataset quality, model generalization, and deployment feasibility in industrial settings. This section categorizes potential threats into internal threats, which are directly related to the research methodology and dataset preparation, and external threats, which arise from practical implementation challenges beyond the controlled research environment.

## 8.1 Construct Validity

Construct validity refers to how well the experimental setup and dataset represent realworld defect detection scenarios. Since SurgScan relies on a curated dataset, it is crucial to ensure that the dataset accurately reflects the variability observed in industrial manufacturing environments. To improve construct validity, the dataset was curated with high-resolution images of actual surgical instruments, ensuring that defect annotations were based on real-world defects rather than synthetic augmentations.

Another critical aspect of construct validity is the classification of non-defective images. To prevent the model from learning artifacts or biases from background objects, a Miscellaneous (Misc) class was introduced. This class contains non-surgical objects and irrelevant background elements, ensuring that the model distinguishes between actual defects and irrelevant visual noise. By training the model on both defect-specific and non-defective instrument images, the likelihood of false positives was significantly reduced.

Despite these measures, construct validity may be impacted by the limited number of manufacturers contributing to the dataset. Future studies should incorporate defect samples from multiple production facilities to further improve dataset generalization and reduce potential biases related to specific manufacturing processes.

## 8.2 Internal Threats

A primary concern in deep learning-based defect detection is dataset bias and class imbalance, where an uneven distribution of defects across different instruments can lead to skewed learning and reduced generalization. If certain defect types are overrepresented, the model may develop a bias toward recognizing frequent defect categories while failing to detect rare defects. To address this, the dataset was carefully curated to ensure that all instruments contained all defect types, preventing the model from favoring common defects while underperforming on subtle or rare anomalies. Additionally, data augmentation techniques such as rotation, brightness adjustments, noise addition, and contrast modifications were applied to improve dataset diversity, minimizing overfitting and enhancing generalization. Despite these measures, data augmentation cannot fully replace naturally occurring variations in defect patterns. Acquiring defect samples from multiple manufacturers and production environments would further enhance real-world adaptability, ensuring that the model generalizes across diverse industrial conditions.

Another major internal challenge is annotation consistency and human error, which can introduce inconsistencies in defect labeling. Surgical instrument defects such as fine scratches, micro-cracks, and early-stage corrosion can be difficult to classify, leading to inter-annotator variability. Annotation reliability was ensured through expert validation using a majority voting approach, minimizing misclassifications before model training. In cases where annotators disagreed, a neutral expert adjudicated the final classification to maintain dataset consistency. Despite these precautions, annotationrelated errors remain a persistent challenge in industrial defect detection, particularly for subtle defects that are difficult to perceive in certain lighting conditions.

Another key internal threat is the presence of non-relevant objects and background noise in images, which could cause the model to misclassify artifacts as defects. Since industrial inspections often involve multiple objects in the frame, it is essential to ensure that the model focuses only on surgical instruments and their defects. To mitigate this issue, a Miscellaneous (Misc) class was introduced, which includes non-surgical objects and irrelevant background elements. Training the model to recognize and ignore non-instrument objects significantly reduced false positive detections, ensuring that only relevant defects were classified.

## 8.3 External Threats

One of the most significant external threats to SurgScan's deployment is variability in real-world lighting conditions and background environments. The dataset was collected under controlled imaging conditions, ensuring consistent lighting and minimal reflections; however, industrial environments introduce unpredictable factors such as shadows, uneven lighting, reflections, and external noise. These environmental factors can influence defect visibility, leading to fluctuations in model performance when deployed in real-world factory settings. To address this challenge, the model was trained with background variations and multiple lighting conditions, improving its adaptability to real-world imaging setups. Additionally, deployment guidelines were formulated to specify optimal lighting conditions for maximum defect detection accuracy, ensuring consistent model performance in industrial settings.

Another external challenge is the hardware dependency of deep learning models. Unlike mobile-based models optimized for edge computing, SurgScan is designed for high-performance industrial processing, requiring substantial computational resources for real-time analysis. While this setup allows for high precision defect detection without compromising accuracy, it may limit adoption in industries with resourceconstrained environments. However, since the framework is intended for industrial inspection settings where high-performance GPUs are available, hardware limitations do not pose a significant constraint. Additionally, the efficiency of YOLOv8 allows for high-speed defect classification, ensuring that the system can process thousands of instruments per day without major computational bottlenecks.

Scalability is another external concern, particularly when integrating the model into automated conveyor-based inspection systems. The current implementation focuses on standalone industrial inspections, but high-volume manufacturing environments require defect detection models that can operate in real-time on continuous production lines. While SurgScan's architecture is designed to be scalable, real-time factory integration requires ensuring that the model maintains its performance in highthroughput environments. Ongoing efforts to optimize inference speed and minimize processing delays will be crucial for seamless industrial deployment.

By systematically addressing internal and external threats, the SurgScan framework ensures high reliability, accuracy, and practical usability for automated defect classification in surgical instrument manufacturing. Dataset biases and annotation inconsistencies were mitigated through expert validation and data augmentation, while environmental variability was accounted for through controlled imaging conditions and adaptive training strategies. Additionally, the model's efficiency and scalability make it well-suited for industrial adoption, ensuring compliance with international quality control standards. By ensuring dataset diversity, robust annotation validation, and adaptable deployment strategies, SurgScan is designed to maintain high reliability in large-scale industrial applications.

# 9 Conclusions

Ensuring the quality of surgical instruments is essential for patient safety and compliance with international medical standards. Traditional manual inspection methods are slow, inconsistent, and costly, making them unsuitable for large-scale industrial applications. To overcome these challenges, we developed SurgScan, a real-time deep-learning framework leveraging the YOLOv8 architecture for automated surgical instrument classification and defect detection.

To address these challenges, we introduced SurgScan, a real-time deep-learning framework leveraging the YOLOv8 architecture for automated surgical instrument classification and defect detection. The experimental results demonstrate that SurgScan achieves state-of-the-art performance, outperforming CNN-based models such as ResNet, ResNext, and EfficientNet in accuracy, precision, recall, and F1-score. Additionally, SurgScan maintains real-time inference speeds, making it well-suited for industrial-scale deployment without compromising accuracy.

Our findings demonstrate that SurgScan achieves superior defect classification, particularly in detecting critical defects such as corrosion, scratches, and structural misalignments, which are often overlooked in manual inspections. The framework effectively balances high detection accuracy with computational efficiency, enabling scalability for real-world industrial applications. The integration of advanced preprocessing techniques, contrast-based enhancements, and extensive dataset augmentation further enhances the model's robustness across diverse imaging conditions. Statistical validation, including Chi-Square and ANOVA tests, reinforces the significance of SurgScan's defect detection capabilities, showing that defect occurrence varies significantly across instrument types and that contrast-enhanced preprocessing improves defect classification accuracy.

A key contribution of this research is the development of a high-resolution, expertannotated dataset for surgical instrument quality control. Comprising 8,573 original images and expanded to over 102,000 through augmentation, it provides a comprehensive benchmark covering 11 instrument types and five major defect categories. This dataset not only enables rigorous evaluation of deep learning models but also serves as a valuable resource for advancing AI-driven defect detection in medical manufacturing. This expert-annotated dataset not only facilitates rigorous evaluation of deep learning models but also serves as a valuable benchmark for future research in automated surgical instrument inspection. This work provides a strong foundation for industrial adoption, offering a cost-effective, scalable, and reliable alternative to traditional quality control approaches.

Despite its high accuracy, SurgScan has limitations. Detecting low-contrast scratches and micro-level imperfections remains challenging under variable lighting conditions. While contrast-based preprocessing enhances defect visibility, certain faint defects may still be misclassified due to subtle texture variations. Additionally, ensuring seamless deployment in high-volume production lines presents integration challenges, particularly for conveyor-based inspection systems. Additionally, real-world deployment in high-volume production lines requires seamless integration with automated manufacturing workflows, where instruments move continuously on conveyor systems.

## 10 Future Work

While SurgScan has demonstrated high accuracy and real-time efficiency in surgical instrument defect detection, several areas warrant further research to enhance its generalization, adaptability, and industrial scalability. Expanding the dataset to include a wider variety of surgical instruments, material compositions, and real-world defect samples will improve the model's robustness and applicability across different manufacturing conditions. Additionally, integrating hybrid deep learning approaches, such as combining transformers with CNNs, could enhance defect localization and classification accuracy, particularly for low-contrast micro-defects.

Another promising avenue is the incorporation of semi-supervised and unsupervised learning techniques, allowing the model to continuously learn from real-world data and improve its performance without requiring extensive manual annotations. Furthermore, leveraging multi-modal imaging technologies—such as infrared, X-ray, or hyperspectral imaging—could facilitate the detection of internal structural defects that are not visible in standard RGB images. Deploying SurgScan in real-world industrial environments will be crucial for evaluating its robustness across different factory setups, lighting conditions, and production workflows. Collaborations with surgical instrument manufacturers and regulatory bodies will also help refine the model to meet ISO and FDA-GMP quality assurance standards, ensuring seamless industry adoption.

Moreover, the integration of Explainable AI (XAI) techniques could enhance the interpretability of defect classifications, providing manufacturers with transparent and actionable insights into the decision-making process. Future improvements should also focus on optimizing inference speed for high-throughput manufacturing environments, ensuring that SurgScan can process thousands of instruments per hour without compromising detection accuracy. By addressing these challenges, SurgScan can evolve into a fully automated, high-precision quality control system, significantly reducing reliance on manual inspections while ensuring compliance with international medical standards.

## References

[1] C.P.M.D.P.M.J.G. W. Waked, A. Simpson, Sterilization wrap inspections do not adequately evaluate instrument sterility. Clinical Orthopaedics and Related Research (2007). https://doi.org/10.1097/BLO.0b013e318065b0bc. URL https://www.semanticscholar.org/paper/59d2063bc47300c301bef2b23d97187eb182ee5b

[2] D.O. Pontes, D.d.M. Costa, P.P. da Silva Pereira, G.S. Whiteley, T. Glasbey, A.F.V. Tipple, Adenosine triphosphate (atp) sampling algorithm for monitoring the cleanliness of surgical instruments. PloS One **18**(8), e0284967 (2023)

[3] F.A.N. Aini, A.Z. Purwalaksana, I.P. Manalu, *Object detection of surgical instruments for assistant robot surgeon using knn*, in *2019 international conference on advanced mechatronics, intelligent manufacture and industrial automation (ICAMIMIA)* (IEEE, 2019), pp. 37–40

[4] X.H. Liu, C.H. Hsieh, J.D. Lee, S.T. Lee, C.T. Wu, *A vision-based surgical instruments classification system*, in *2014 International Conference on Advanced Robotics and Intelligent Systems (ARIS)* (IEEE, 2014), pp. 72–77

[5] S. Bodenstedt, A. Ohnemus, D. Katic, A.L. Wekerle, M. Wagner, H. Kenngott, B. Müller-Stich, R. Dillmann, S. Speidel, Real-time image-based instrument classification for laparoscopic surgery. arXiv preprint arXiv:1808.00178 (2018)

[6] M. Xue, L. Gu, *Surgical instrument segmentation method based on improved MobileNetV2 network*, in *2021 6th International Symposium on Computer and Information Processing Technology (ISCIPT)* (IEEE, 2021), pp. 744–747

[7] A.O. Campos-Montes, C.J. Vega-Urquizo, M.G.S. Paredes, K. Acuna-Condori, *Implementation of a computer-assisted surgical instrument sterilization system based on deep learning for health centers*, in *2022 IEEE ANDESCON* (IEEE, 2022), pp. 1–6

[8] S.A. Haider, O.A. Ho, S. Borna, C.A. Gomez-Cabello, S.M. Pressman, D. Cole, A. Sehgal, B.C. Leibovich, A.J. Forte, Use of multimodal artificial intelligence in surgical instrument recognition. Bioengineering **12**(1), 72 (2025)

[9] H.B. Le, T.D. Kim, M.H. Ha, A.L.Q. Tran, D.T. Nguyen, X.M. Dinh, *Robust Surgical Tool Detection in Laparoscopic Surgery using YOLOv8 Model*, in *2023 International Conference on System Science and Engineering (ICSSE)* (IEEE, 2023), pp. 537–542

[10] B. Ran, B. Huang, S. Liang, Y. Hou, Surgical instrument detection algorithm based on improved yolov7x. Sensors **23**(11), 5037 (2023)

[11] K. Lam, F.P.W. Lo, Y. An, A. Darzi, J.M. Kinross, S. Purkayastha, B. Lo, Deep learning for instrument detection and assessment of operative skill in surgical videos. IEEE Transactions on Medical Robotics and Bionics **4**(4), 1068–1071 (2022)

[12] S. Muller, F. Despinoy, D. Bratbak, E. Tronvik, P. Jannin, *Real-time phase recognition in novel needle-based intervention: a multi-operator feasibility study*, in *Medical Imaging 2017: Image-Guided Procedures, Robotic Interventions, and Modeling*, vol. 10135 (SPIE, 2017), pp. 115–121

[13] L. Sun, X. Chen, Pixel-wise contrastive learning for multi-class instrument segmentation in endoscopic robotic surgery videos using dataset-wide sample queues. IEEE Access (2024)

[14] K. Lam, F.P.W. Lo, Y. An, A. Darzi, J.M. Kinross, S. Purkayastha, B. Lo, Deep learning for instrument detection and assessment of operative skill in surgical videos. IEEE Transactions on Medical Robotics and Bionics **4**(4), 1068–1071 (2022)

[15] E.D. Dominguez, B. Rocos, Patient safety incidents caused by poor quality surgical instruments. Cureus **11**(6) (2019)

[16] T. Brophy, P. Srodon, C. Briggs, P. Barry, J. Steatham, M. Birch, Quality of surgical instruments. The Annals of The Royal College of Surgeons of England **88**(4), 390–393 (2006)

[17] S. Wang, X. Yin, B. Ge, Y. Gao, H. Xie, L. Han, *Machine Vision for Automated Inspection of Surgical Instruments*, in *2009 3rd International Conference on Bioinformatics and Biomedical Engineering* (2009), pp. 1–4. https://doi.org/10.1109/ICBBE.2009.5163022

[18] H. Al Hajj, M. Lamard, P.H. Conze, B. Cochener, G. Quellec, Monitoring tool usage in surgery videos using boosted convolutional and recurrent neural networks. Medical image analysis **47**, 203–218 (2018)

[19] M. Alsheakhali, M. Yigitsoy, A. Eslami, N. Navab, *Surgical tool detection and tracking in retinal microsurgery*, in *Medical Imaging 2015: Image-Guided Procedures, Robotic Interventions, and Modeling*, vol. 9415 (SPIE, 2015), pp. 245–250

[20] B. Choi, K. Jo, S. Choi, J. Choi, *Surgical-tools detection based on Convolutional Neural Network in laparoscopic robot-assisted surgery*, in *2017 39th annual international conference of the IEEE engineering in medicine and biology society (EMBC)* (Ieee, 2017), pp. 1756–1759

[21] C. Gao, M. Unberath, R. Taylor, M. Armand, Localizing dexterous surgical tools in x-ray for image-based navigation. arXiv preprint arXiv:1901.06672 (2019)

[22] K. He, G. Gkioxari, P. Dollár, R. Girshick, *Mask r-cnn*, in *Proceedings of the IEEE international conference on computer vision* (2017), pp. 2961–2969

[23] D. Hendrycks, K. Gimpel, Gaussian error linear units (gelus). arXiv preprint arXiv:1606.08415 (2016)

[24] J. Hu, L. Shen, G. Sun, *Squeeze-and-excitation networks*, in *Proceedings of the IEEE conference on computer vision and pattern recognition* (2018), pp. 7132–7141

[25] Y. Jin, Q. Dou, H. Chen, L. Yu, J. Qin, C.W. Fu, P.A. Heng, Sv-rcnet: workflow recognition from surgical videos using recurrent convolutional network. IEEE transactions on medical imaging **37**(5), 1114–1126 (2017)

[26] Y. Jin, H. Li, Q. Dou, H. Chen, J. Qin, C.W. Fu, P.A. Heng, Multi-task recurrent convolutional network with correlation loss for surgical video analysis. Medical image analysis **59**, 101572 (2020)

[27] L. Joskowicz, C. Milgrom, A. Simkin, L. Tockus, Z. Yaniv, Fracas: a system for computer-aided image-guided long bone fracture surgery. Computer Aided Surgery: Official Journal of the International Society for Computer Aided Surgery (ISCAS) **3**(6), 271–288 (1998)

[28] A. Krupa, J. Gangloff, C. Doignon, M.F. De Mathelin, G. Morel, J. Leroy, L. Soler, J. Marescaux, Autonomous 3-d positioning of surgical instruments in robotized laparoscopic surgery using visual servoing. IEEE transactions on robotics and automation **19**(5), 842–853 (2003)

[29] T. Kurmann, P. Marquez Neila, X. Du, P. Fua, D. Stoyanov, S. Wolf, R. Sznitman, *Simultaneous recognition and pose estimation of instruments in minimally invasive surgery*, in *Medical Image Computing and Computer-Assisted InterventionMICCAI 2017: 20th International Conference, Quebec City, QC, Canada, September 11-13, 2017, Proceedings, Part II 20* (Springer, 2017), pp. 505–513

[30] I. Laina, N. Rieke, C. Rupprecht, J.P. Vizca´ıno, A. Eslami, F. Tombari, N. Navab, *Concurrent segmentation and localization for tracking of surgical instruments*, in *Medical Image Computing and Computer-Assisted Intervention- MICCAI 2017: 20th International Conference, Quebec City, QC, Canada, September 11-13, 2017, Proceedings, Part II 20* (Springer, 2017), pp. 664–672

[31] Y. Liu, Z. Zhao, F. Chang, S. Hu, An anchor-free convolutional neural network for real-time surgical tool detection in robot-assisted surgery. IEEE Access **8**, 78193–78201 (2020)

[32] K. Mishra, R. Sathish, D. Sheet, *Learning latent temporal connectionism of deep residual visual abstractions for identifying surgical tools in laparoscopy procedures*, in *Proceedings of the IEEE Conference on Computer Vision and Pattern Recognition Workshops* (2017), pp. 58–65

[33] T.K. Kim, Understanding one-way anova using conceptual figures. Korean journal of anesthesiology **70**(1), 22 (2017)

[34] M.L. McHugh, The chi-square test of independence. Biochemia medica **23**(2), 143–149 (2013)